\documentclass[11pt,a4paper]{article}
\usepackage[hyperref]{acl2020}

\aclfinalcopy

\usepackage{url}
\usepackage{times}
\usepackage{latexsym}

\usepackage{url}
\usepackage{proof}
\usepackage{xspace}
\usepackage{multirow}
\usepackage{forest}
\usepackage{caption}
\usepackage{subcaption}
\usepackage{rotating}
\usepackage{dsfont}
\usepackage{amsmath}
\usepackage{amssymb}
\usepackage{amsthm}
\usepackage{algorithm}
\usepackage{float}
\usepackage{tikz}
\usepackage{fontawesome}
\usetikzlibrary{plotmarks}
\newcommand\marksymbol[2]{\tikz[#2,scale=1.2]\pgfuseplotmark{#1};}

\newcommand{\namecite}[1]{\newcite{#1}}
\usepackage{bm}
\newcommand{\pad}{\ensuremath{\langle\text{pad}\rangle}\xspace}

\newcommand{\tuple}[1]{\ensuremath{\langle {#1} \rangle}}

\newcommand{\notes}[1]{}

\newcommand{\nextfunc}{\operatornamewithlimits{\mathrm{next}}}

\theoremstyle{definition}

\theoremstyle{plain}

\newcommand{\vech}{\ensuremath{\mathbf{h}}}

\newcommand{\ith}[1]{\ensuremath{i^{{th}}}}

\newcount\permx
\newcount\permy
\def\permdot#1#2{
\permx=#1 \advance\permx by-1
\permy=#2 \advance\permy by-1
\psframe[fillcolor=black, fillstyle=solid]
(\permx,\permy)(#1, #2)
}

\newcommand{\argmax}{\operatornamewithlimits{\mathbf{argmax}}}

\newcommand{\toptop}{\operatornamewithlimits{\mathbf{top}}}

\newcommand{\boxnum}[1]{{\setlength{\fboxsep}{1pt}\raisebox{1pt}{\hspace{1pt}\fbox{\tiny #1}\hspace{1pt}}}}
\newcommand{\ind}[1]{\ensuremath{_{\kern-0.5pt\boxnum{#1}}}}

\newcommand{\vecx}{\ensuremath{\bm{x}}\xspace}
\newcommand{\vecy}{\ensuremath{\bm{y}}\xspace}

\def\namecite{\newcite}

\newcommand{\smallnt}[1]{\ensuremath{_{\mbox{\tiny PP}}}\xspace}

\newcommand{\pseudocode}{Algorithm}
\floatname{algorithm}{\pseudocode}

\iffalse

\else

\fi

\newcommand{\eos}{\mbox{\scriptsize \texttt{<eos>}}\xspace}

\newcommand{\RAL}{\ensuremath{\mathrm{RAL}}\xspace}

\definecolor{chocolate}{rgb}{0.28, 0.02, 0.03}
\definecolor{PaleGreen}{rgb}{0.33, 0.545,0.33}
\definecolor{colorC0}{RGB}{51,113, 169}
\definecolor{colorC1}{RGB}{243,130,37}
\usepackage{array}
\usepackage{diagbox}

\definecolor{dollarbill}{rgb}{0.52, 0.73, 0.4}
\definecolor{deepmagenta}{rgb}{0.8, 0.0, 0.8}
\definecolor{coralred}{rgb}{1.0, 0.25, 0.25}

\title{Opportunistic Decoding with Timely Correction \\for Simultaneous Translation}

\author{Renjie Zheng $^{1,2,}$\thanks{\; These authors contributed equally.} \quad
  Mingbo Ma $^{2, \ast}$ \quad
  Baigong Zheng $^{2}$ \quad
  Kaibo Liu$^{2}$ \quad
  Liang Huang $^{1,2}$
\\
  $^{1}$Oregon State University, Corvallis, OR, USA \\
  $^{2}$Baidu Research, Sunnyvale, CA, USA \\
  \texttt{\{renjiezheng,mingboma\}@baidu.com} \\
}

\date{}

\begin{document}
\maketitle
\begin{abstract}
Simultaneous translation has many important 
application scenarios and 
attracts much attention from both academia
and industry recently. 
Most existing frameworks, however, have difficulties 
in balancing between the translation quality and latency,
i.e., 
the decoding policy is usually 
either too aggressive or too conservative.
We propose an opportunistic decoding technique with
timely correction ability,
which always (over-)generates
a certain mount of extra words
at each step to keep the audience on track 
with the latest information. 
At the same time,
it also corrects, in a timely fashion,
the mistakes in the former overgenerated words
when observing more source context
to ensure high translation quality.
Experiments show our technique achieves 
substantial reduction in latency and up to +3.1 increase in BLEU, with 
revision rate under 8\% in Chinese-to-English and 
English-to-Chinese translation. 
\end{abstract}

\section{Introduction}
Simultaneous translation, which starts translation 
before the speaker finishes, 
is extremely useful in many scenarios, 
such as international conferences, travels, 
and so on.
In order to achieve low latency, 
it is often inevitable to generate
target words 
with insufficient source information, which makes this task
extremely challenging.

Recently, there are many efforts 
towards balancing the translation latency and quality 
with mainly two types of approaches.
On one hand,
\namecite{ma+:2019} 
propose very simple 
frameworks that 
decode 
following a {\em fixed-latency policy} such as wait-$k$.
On the other hand, 
there are many attempts to learn an {\em{adaptive policy}}
which enables the model to decide {\sc{read}} or 
{\sc{write}} action on the fly 
using various techniques such as 
reinforcement learning \cite{Gu+:2017,ashkan+:2018,grissom+:2014}, 
supervised learning over pseudo-oracles \cite{zheng+:2019b}, 
imitation learning \cite{zheng+:2019}, model ensemble \cite{zheng+:2020}
or monotonic attention \cite{xtma+:2019,Arivazhagan+:2019}.

\begin{figure}[t]
\centering
\includegraphics[width=7cm]{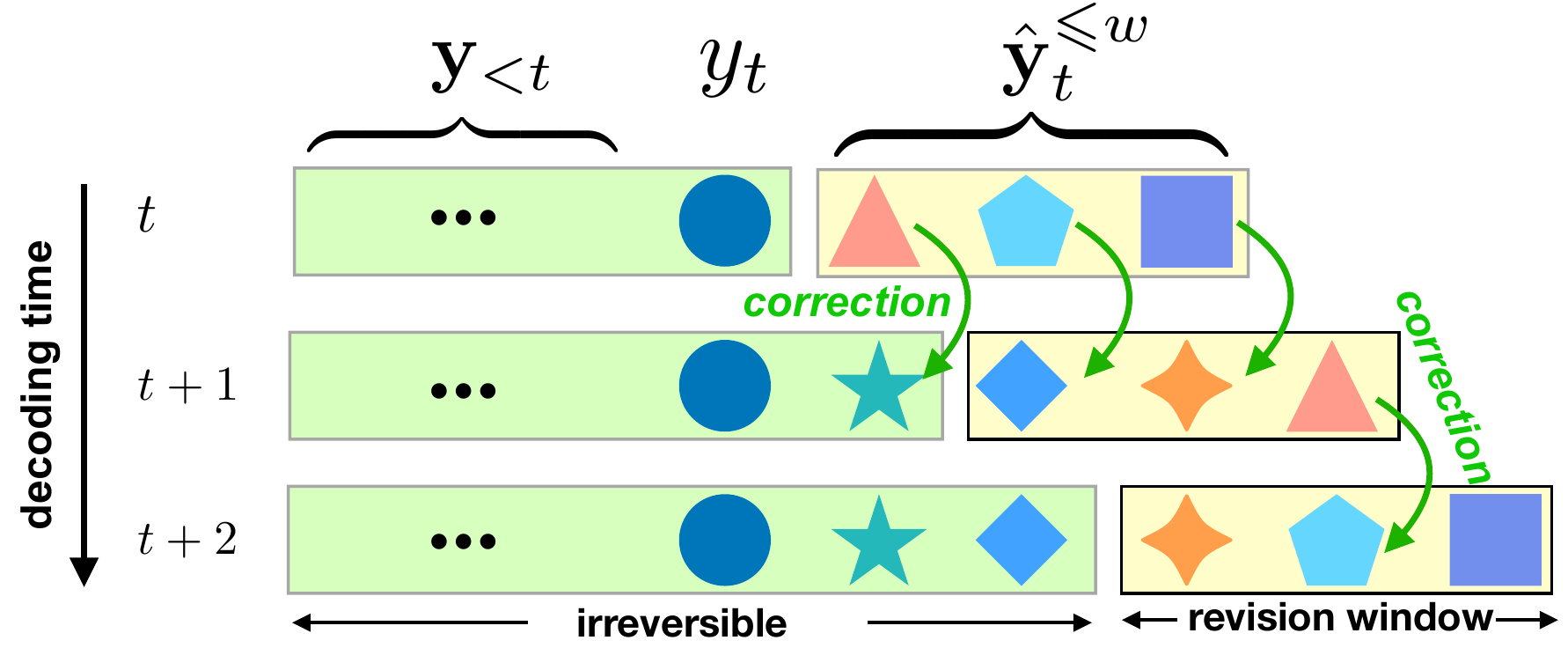}
\captionof{figure}{Besides $y_t$, opportunistic decoding continues to
generate additional $w$ words which are represented as $\hat{\vecy}_{t}^{\leqslant w}$. The timely correction only
revises this part in future steps. Different shapes denote different words.
In this example, from step $t$ to $t+1$, all previously opportunistically decoded words are revised, and
an extra triangle word is generated in opportunistic window.
From step $t+1$ to $t+2$, two words from previous opportunistic window are kept and only the triangle word is revised.}
\label{fig:intro}
\vspace{-0.5cm}
\end{figure}

\if 
Though above efforts improve 
both latency and translation quality,
the translation model still suffers from the 
predetermined policy which is obtained either from
experience (fixed policy) nor 
training data (adaptive policy)
and ignore the individual characteristics of 
each sentence during inference time case by case.
During inference time, a conservative policy makes
unnecessary pauses which delays 

Even for the so-called adaptive policy-based methods,
the model parameters are still learned based on 
certain predefined, fixed hyper-parameters (or threshold)
which are decided by balancing the latency and accuracy
that are compromised over the entire training or validation data.
Especially when we apply the learned policy to a different
domain, the chosen policy is usually too aggressive or
conservative. 
\fi

Though the existing efforts improve the performance in
both translation latency and quality 
with more powerful frameworks,
it is still difficult to choose an appropriate policy
to explore the optimal balance between latency and quality
in practice, especially 
when the policy is trained and 
applied in different domains.
Furthermore, all existing approaches are incapable of 
correcting the mistakes from previous steps.
When the former steps commit errors, 
they will be propagated to the later steps,
inducing more mistakes to the future.

Inspired by our previous work on speculative beam search
\cite{zheng2019speculative}, we propose an opportunistic decoding technique with timely correction mechanism to address the 
above problems.
As shown in Fig.~\ref{fig:intro},
our proposed method always decodes more words than
the original policy at each step 
to catch  up with the speaker and reduce the latency.
At the same time,
it also employs a timely correction mechanism to
review the extra outputs
from previous steps with more source context, 
and revises these outputs with current preference
when there is a disagreement.
Our algorithm can be used in both speech-to-text 
and speech-to-speech simultaneous translation \cite{oda+:2014,bangalore+:2012,mahsa+:2013}.
In the former case, 
the audience will not be overwhelmed by the modifications
since we only review and modify the last 
few output words with a relatively low revision rate.
In the later case, 
the revisable extra words can be used in look-ahead window
in incremental TTS \cite{ma2019incremental}.
By contrast, the alternative re-translation strategy \cite{arivazhagan2020re}
will cause non-local revisions which
makes it impossible to be used in incremental TTS.

\if
We always prefer the outputs with more context over
those with less context. 
Thanks to the timely correction ability,
during decoding time, 
it becomes appropriate to choose more aggressive policy 
to reduce latency without sacrificing the accuracy.
\fi

We also define, for the first time, two metrics for revision-enabled simultaneous translation:
a more general latency metric {\em Revision-aware 
Average Lagging} (RAL)
as well as the {\em revision rate}. 
We demonstrate the effectiveness of our proposed technique
using
 fixed \cite{ma+:2019} and 
adaptive \cite{zheng+:2019b} policies
in both Chinese-to-English and English-to-Chinese 
translation.

  \vspace{-5pt}
\section{Preliminaries}

\paragraph{Full-sentence NMT.}
The conventional full-sentence NMT processes 
the source sentence $\vecx = (x_1,...,x_n)$
with an encoder, where $x_i$ represents 
an input token.
The decoder on the target side (greedily)
selects the highest-scoring word $y_t$ given 
source representation $\vech$ and previously 
generated target tokens, $\vecy_{<t}=(y_1,...,y_{t-1})$,
and the final hypothesis $\vecy = (y_1,...,y_t)$ with $y_t = \eos$
has the highest probability:
\vspace{-0.1cm}
\begin{equation}
\vspace{-0.1cm}
p(\vecy \mid \vecx) = \textstyle\prod_{t=1}^{|\vecy|}  p(y_t \mid \vecx,\, \vecy_{<t})
\label{eq:gensentscore}
\vspace{-0cm}
\end{equation}

\paragraph{Simultaneous Translation.}
Without loss of generality, regardless the 
actual design of policy,
simultaneous translation is represented as:
\vspace{-0.1cm}
\begin{equation}
\vspace{-0.1cm}
p_g(\vecy \mid \vecx) = \textstyle\prod_{t=1}^{|\vecy|}  p(y_t \mid \vecx_{\leqslant g(t)},\, \vecy_{<t})
\label{eq:gensentscore2}
\vspace{-0cm}
\end{equation}
where $g(t)$ can be used to represent any arbitrary fixed or 
adaptive policy.
For simplicity,
we assume the policy is given and does not distinguish
the difference between two types of policies.

  \vspace{-5pt}
\section{Opportunistic Decoding with Timely Correction and Beam Search}

\begin{figure*}[t]
\centering
\includegraphics[width=12cm]{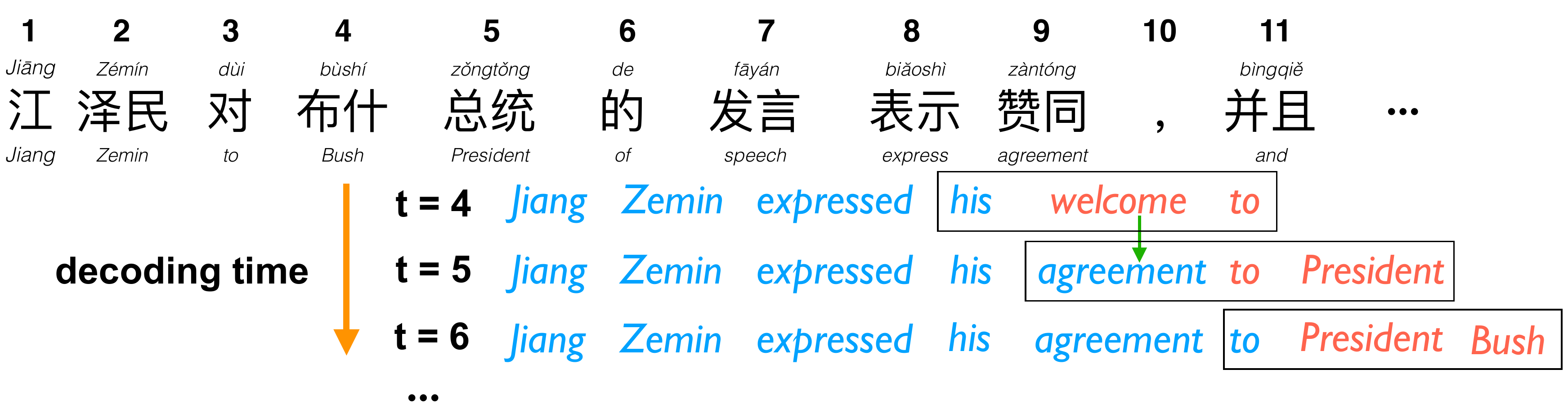}
\captionof{figure}{The decoder generates target word $y_4=$ ``his'' and two extra words 
``welcome to'' at step $t=4$ when input $x_9=$ 
``z\`{a}nt\'{o}ng'' (``agreement'') is not available yet. When the model receives  
$x_9$ 
at step $t=5$, the decoder
immediately corrects the previously made mistake ``welcome'' with ``agreement'' and emits two additional target words (``to President'').
The decoder not only is capable to fix the previous mistake, but also has enough information to perform more correct generations.
Our framework benefits from 
opportunistic decoding with reduced latency here.
Note though the word ``to'' is generated in step $t=4$, it only becomes
irreversible at step $t=6$.}
\label{fig:decoding}
\vspace{-0.5cm}
\end{figure*}


\paragraph{Opportunistic Decoding.}
For simplicity, we first apply this method to fixed policies.
We define the original decoded word sequence
at time step $t$ with $y_t$, which represents the word
that is decoded in time step $t$ with original model.
We denote the additional decoded words at time step $t$ 
as $\hat{\vecy}_t^{\leqslant  w}=(y_t^1,...,y_t^w)$, where 
$w$ denote the number of extra decoded words.
In our setting, 
the decoding process is as follows:
\vspace{-0.1cm}
\begin{equation}
\vspace{-0.1cm}
\begin{split}
&p_g(y_t \circ \hat{\vecy}_{t}^{\leqslant  w} \mid \vecx_{\leqslant g(t)}) = \\
&p_g(y_{t} \mid \vecx_{\leqslant g(t)})\textstyle\prod_{i=1}^{w}  
p_g(\hat{y}_t^i \mid \vecx_{\leqslant g(t)},\, y_{t}\circ \hat{\vecy}_{t}^{<i})
\end{split}
\end{equation}
where $\circ$ is the string concatenation operator.

We treat the procedure for generating 
the extra decoded sequence as 
opportunistic decoding, which prefers to generate
more tokens based on current context.
When we have enough information, this opportunistic decoding
eliminates unnecessary latency and keep the audience on track.
With a certain chance, 
when the opportunistic decoding tends 
to aggressive and generates
inappropriate tokens,
we need to fix the inaccurate token immediately.

\paragraph{Timely Correction.}
In order to deliver the correct information to
the audience promptly and fix
previous mistakes as soon as possible,
we also need to review and modify the previous outputs.

At step $t+1$, when encoder obtains more 
information 
from $\vecx_{\leqslant g(t)}$ to $\vecx_{\leqslant g(t+1)}$, 
the decoder is capable to generate more appropriate candidates
and may revise and replace the 
previous outputs from opportunistic decoding.
More precisely, 
$\hat{\vecy}_{t}^{\leqslant w}$ and 
$y_{t+1} \circ \hat{\vecy}_{t+1}^{\leqslant w-1}$
are two different hypothesis over the same time chunk.
When there is a disagreement,
our model always uses the hypothesis from later step to
replace the previous commits.
Note our model does not change any 
word in $y_{t}$ from previous step
and it only revise the words in $\hat{\vecy}_{t}^{\leqslant w}$.

\paragraph{Modification for Adaptive Policy.} For adaptive policies,
the only difference is, instead of committing a single word,
the model is capable of generating multiple irreversible words.
Thus our proposed methods can be easily applied to adaptive policies.

\paragraph{Correction with Beam Search.}

When the model is committing more than one word at a time, 
we can use beam search to further improve the translation 
quality and reduce revision rate \cite{murray+chiang:2018, ma2019learning}.

The decoder maintains a beam $B_t^k$ of size $b$ at step $t$,
which is  
ordered list of pairs $\tuple{\text{hypothesis, probability}}$,
where $k$ denotes the $k^{th}$ step in beam search.
At each step,
there is an initial beam $B_t^0=[\tuple{\vecy_{t-1}, 1}]$.
We denote one-step
transition from the previous beam to the next as 
\vspace{-0.1cm}
\begin{equation}
\vspace{-0.1cm}
\begin{split}
 & \textstyle B_t^{k+1} = \nextfunc_1^b(B_t^k) \\
 & = \toptop^b  
  \{\tuple{\vecy'\!\circ v, \ u\!\cdot\! p(v | \vecx_{\leqslant g(t)}, \vecy')} \mid \tuple{\vecy', u} \!\in\! B_t^k \} \notag
\end{split}
\end{equation}
where $\toptop^b (\cdot)$ returns the top-scoring $b$ pairs.
Note we do not distinguish the revisable and non-revisable 
output in $\vecy'$ for simplicity.
We also define the multi-step advance beam search function
with recursive fashion as follows:
\vspace{-0.1cm}
\begin{equation}
\vspace{-0.1cm}
 \textstyle \nextfunc^b_i(B_t^k)\! = \!\nextfunc^b_{1}(\nextfunc^b_{i-1}(B_t^k)) \notag
\end{equation}

When the opportunistic decoding window is $w$ at 
decoding step $t$, we define
the beam search over $w+1$ (include the original output)
as follows:
\vspace{-0.1cm}
\begin{equation}
\vspace{-0.1cm}
\textstyle \tuple{{\vecy'_t},u_t} = \textstyle{\toptop^1} \bigl(\nextfunc^b_{n+w}(B_t^0)\bigr)
\label{eq:beamsearchfix}
\end{equation}
where $\nextfunc_{n+w}^b(\cdot)$ performs a beam search
with $n+w$ steps, and generate ${\vecy'_t}$ as the 
outputs which include both original and opportunistic 
decoded words.
$n$ represents the length of $\vecy_t$



  \vspace{-5pt}
\section{Revision-aware AL and Revision Rate}
We define, for the first time, two metrics for revision-enabled 
simultaneous translation.

\subsection{Revision-aware AL}

AL is introduced in \cite{ma+:2019} to measure the average delay
for simultaneous translation.
Besides the limitations that are mentioned in 
\cite{cherry2019thinking}, AL is also 
not sensitive to the modifications to the 
committed words.
Furthermore, in the case of re-translation, 
AL is incapable to measure the meaningful latency anymore.

\begin{figure}[t]
\centering
\includegraphics[width=5cm]{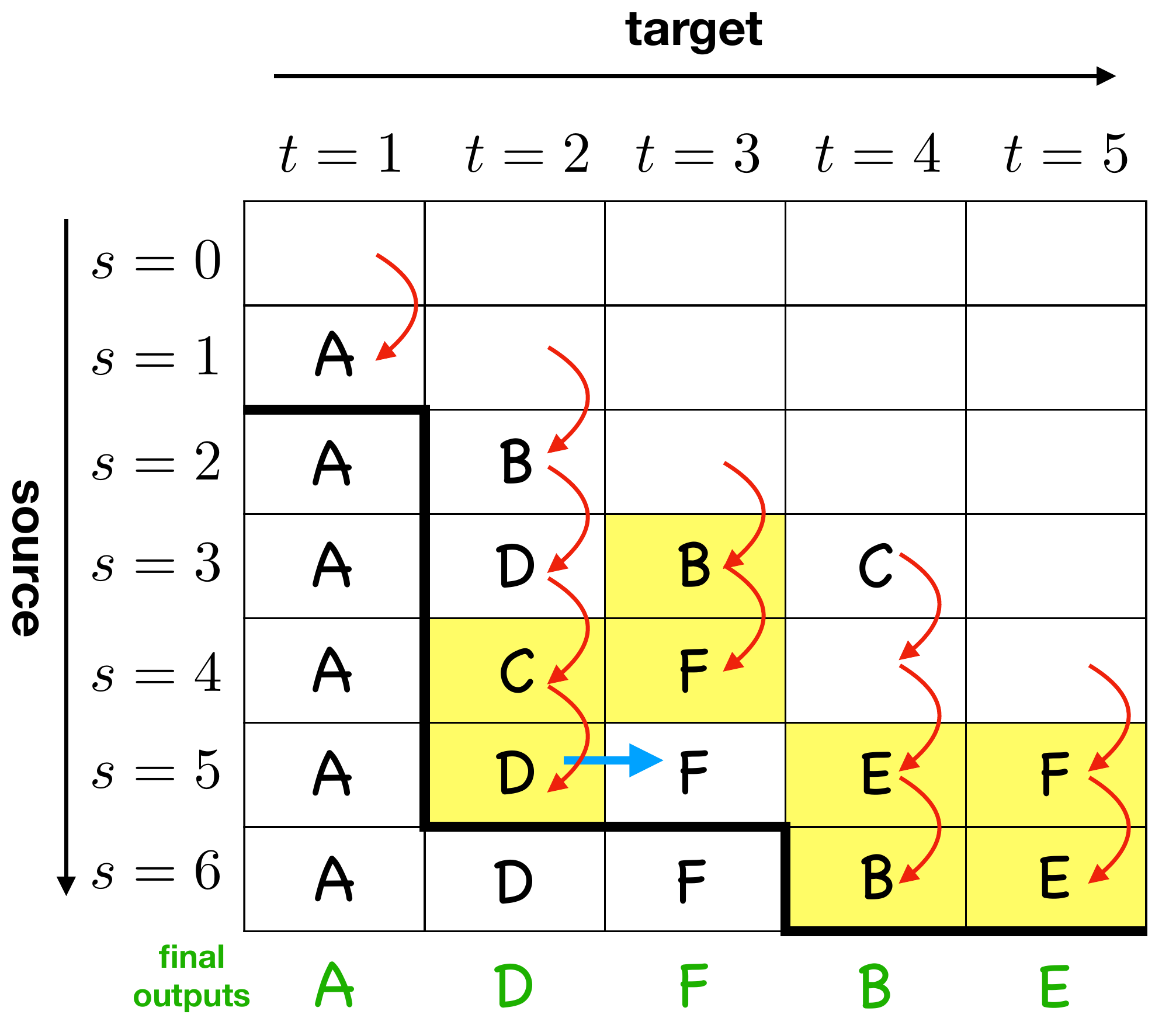}
\vspace{-0.3cm}
\captionof{figure}{The red arrows represent the changes between two 
different commits, and the last changes for each output word is highlighted
with yellow.}
\label{fig:rr}
\vspace{-0.5cm}
\end{figure}

We hereby propose a new latency, Revision-aware AL (RAL), which
can be applied to any kind of translation scenarios, i.e., 
full-sentence translation, use re-translation as simultaneous translation,
fixed and adaptive policy simultaneous translation.
Note that for latency and revision rate calculation,
we count the target side difference respect to the growth of source side.
As it is shown in Fig.~\ref{fig:rr},
there might be multiple changes for each output words during 
the translation, and we only start to calculate the latency for this word
once it agrees with the final results.
Therefore, it is necessary to locate the last change for each word.
For a given source side time $s$, we denote the $t^{th}$ 
outputs on target side
as $f(x_{\leqslant s})_t$.
Then we are able to find the Last Revision (LR) for the $t^{th}$ word on target
side as follows:
\vspace{-0.1cm}
\begin{equation}
\vspace{-0.1cm}
\begin{split}
{LR(t)} = \displaystyle \argmax_{s<|x|} \big(f(x_{\leqslant (s-1)})_t \neq f(x_{\leqslant s})_t \big)\notag
\label{eq:lastrevision}
\end{split}
\end{equation}
From the audience point of view, once the former words are changed,
the audience also needs to take the efforts to read the following as well.
Then we also penalize the later words even there are no changes, 
which is shown with blue arrow in Fig.~\ref{fig:rr}.
We then re-formulate the $\overline{LR}(t)$ as follows (assume $\overline{LR} (0)= 0$):

\begin{figure*}[!h]
\vspace{-6pt}
\begin{tabular}{cc}
\centering
  \includegraphics[width=6.9cm]{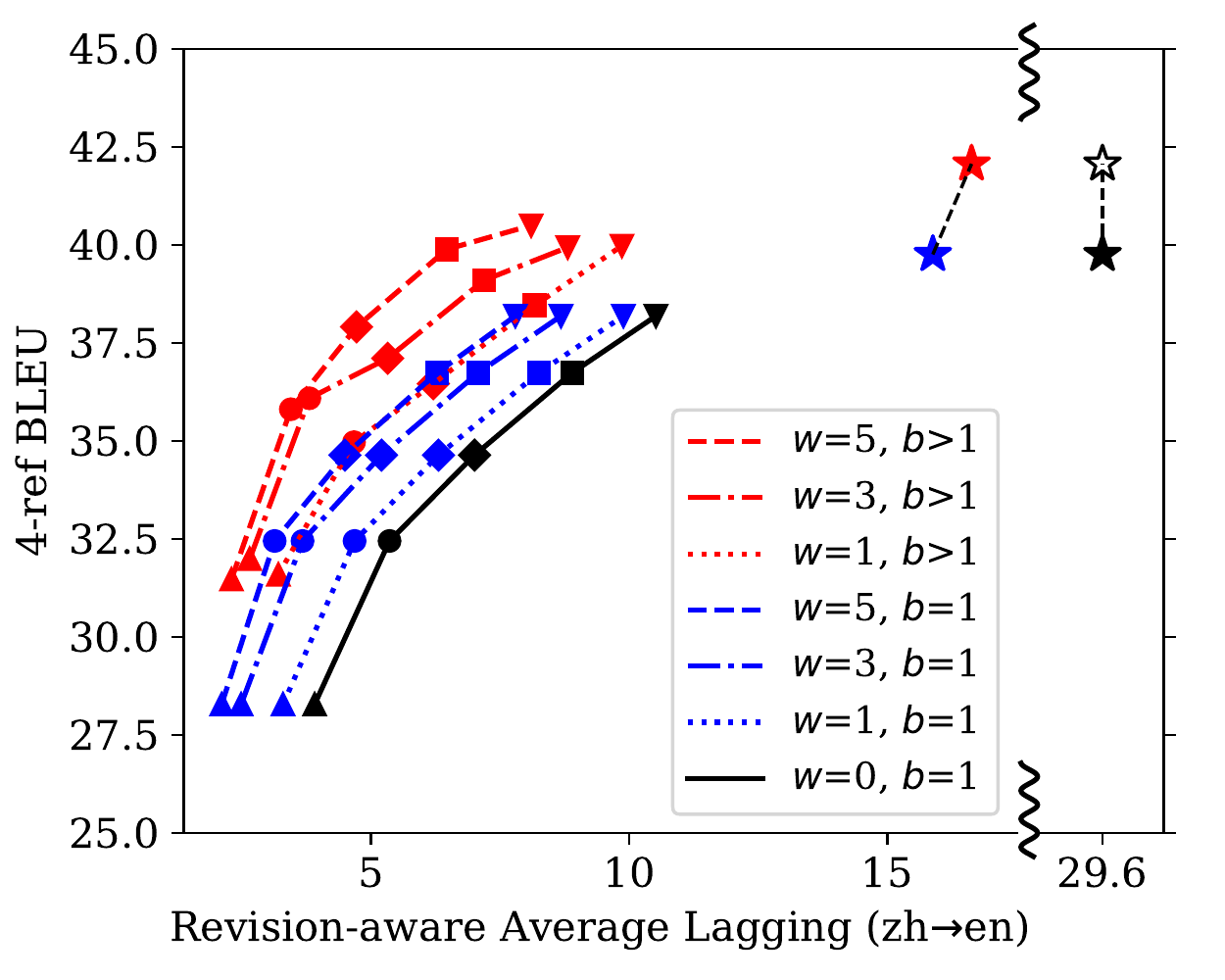} & \qquad
\centering
  \includegraphics[width=6.9cm]{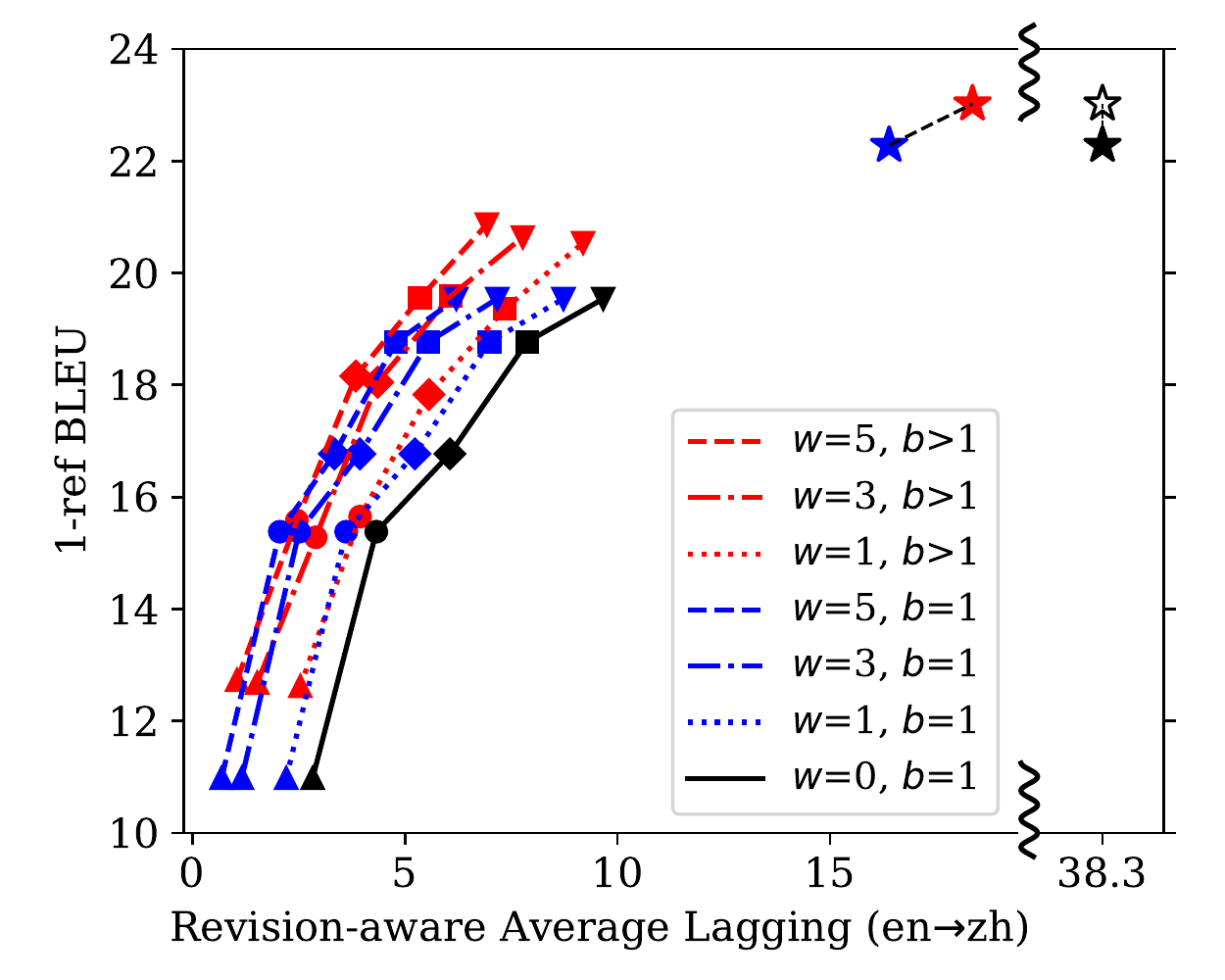}
\end{tabular}\\[-0.4cm]
\caption{
  BLEU against RAL 
  using wait-$k$ polocies.
\textcolor{coralred}{$\blacktriangle$} \textcolor{blue}{$\blacktriangle$} \textcolor{black}{$\blacktriangle$}
: wait-$1$ policies,
\marksymbol{*}{coralred} 
\marksymbol{*}{blue} 
\marksymbol{*}{black} : wait-$3$ policies,
\marksymbol{diamond*}{coralred} 
\marksymbol{diamond*}{blue} 
\marksymbol{diamond*}{black} : wait-$5$ policies,
\marksymbol{square*}{coralred} 
\marksymbol{square*}{blue} 
\marksymbol{square*}{black} : wait-$7$ policies,
\textcolor{coralred}{$\blacktriangledown$} \textcolor{blue}{$\blacktriangledown$}
\textcolor{black}{$\blacktriangledown$}: wait-$9$ policies,\textcolor{blue}{$\bigstar$}
(\textcolor{coralred}{$\bigstar$}): re-translation with pre-trained NMT model
with greedy (beam search) decoding,
\textcolor{black}{$\bigstar$} (\textcolor{black}{\faStarO}): full-sentence translation with pre-trained NMT model with greedy (beam search) decoding.
The baseline for wait-$k$ policies is decoding with $w=0, b=1$.
  \vspace{-10pt}
}
\label{fig:waitk_bleu}
\end{figure*}

\vspace{-10pt}
\begin{figure*}[!h]
\begin{tabular}{cc}
\centering
  \includegraphics[width=6.9cm]{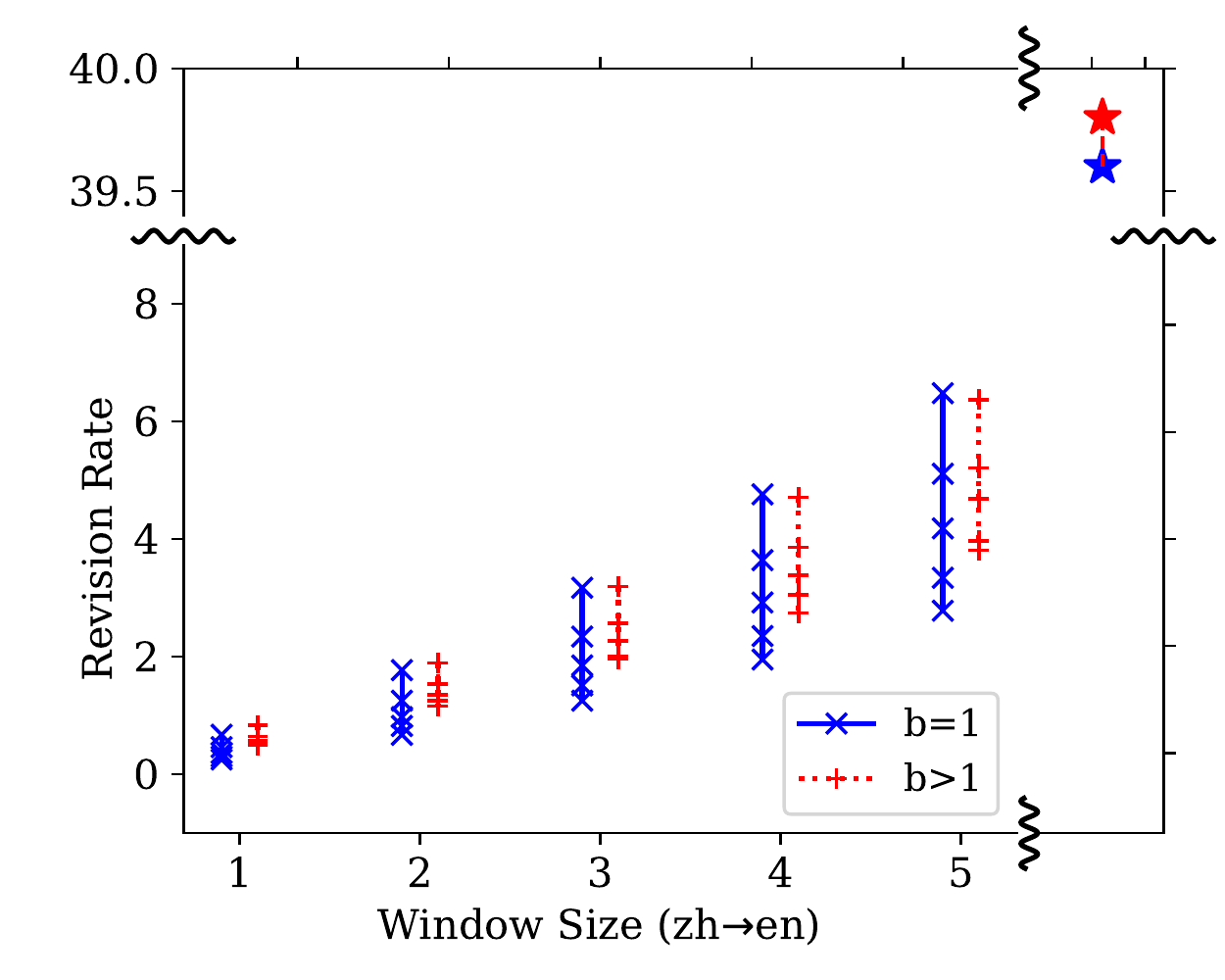} & \qquad
\centering
  \includegraphics[width=6.9cm]{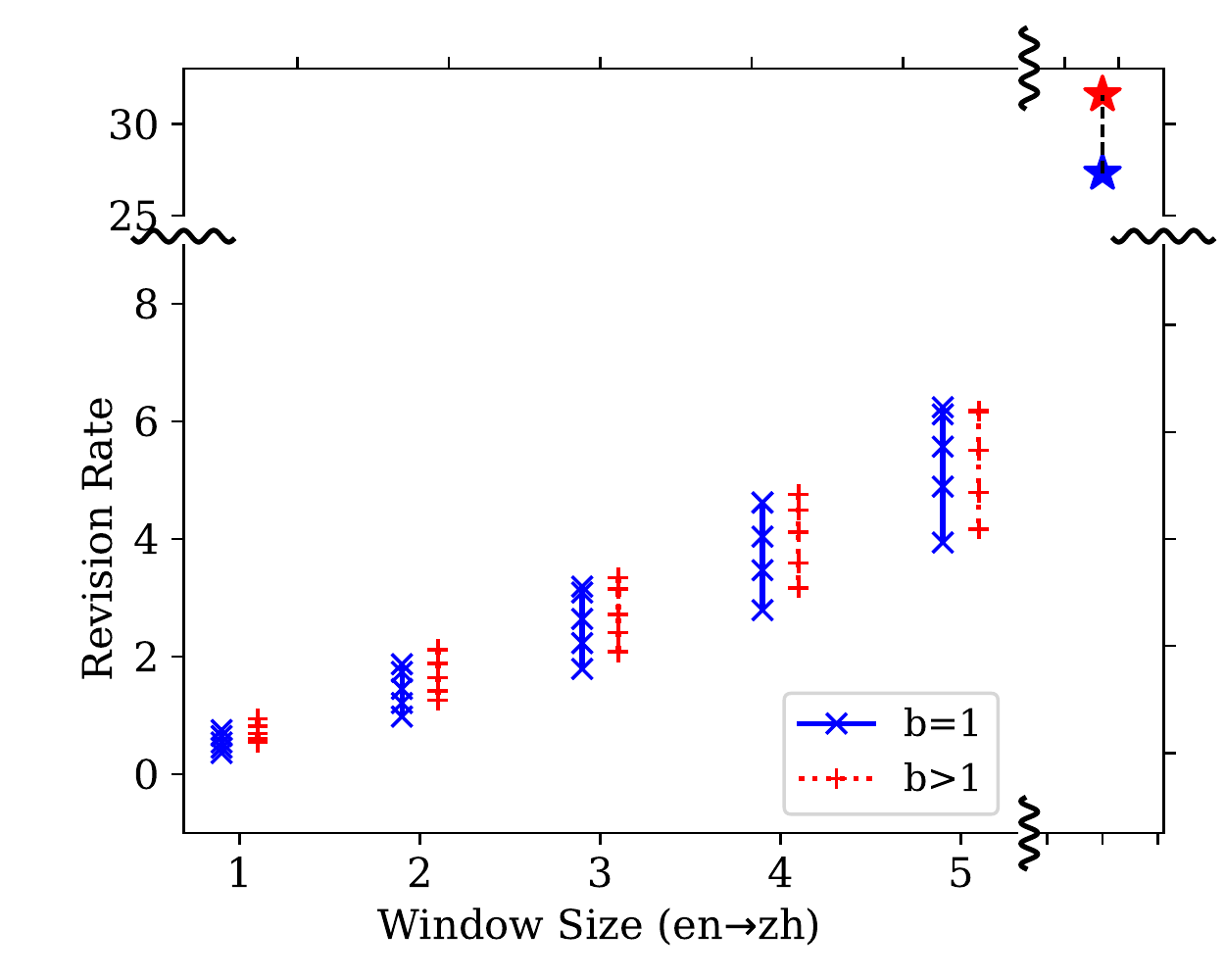}
\end{tabular}\\[-0.4cm]
\captionof{figure}{
Revision rate against window size 
with different wait-$k$ policies.
\textcolor{blue}{$\bigstar$}
(\textcolor{coralred}{$\bigstar$}): re-translation with pre-trained NMT model
with greedy (beam search) decoding.
\vspace{-10pt}
}
\label{fig:waitk_rate}
\end{figure*}

\vspace{-0.1cm}
\begin{equation}
\overline{LR}(t) = max\{\overline{LR}(t-1), {LR(t)} \}
\label{eq:RAL}
\vspace{-0.2cm}
\end{equation}
The above definition can be visualized as the thick black line in Fig.~\ref{fig:rr}.
Similar with original AL, our proposed RAL is defined as follows:
\vspace{-0.1cm}
\begin{equation}
\vspace{-0.1cm}
\RAL(\vecx,\vecy) = \frac{1}{\tau(|\vecx|)} \sum_{t = 1}^{\tau(|\vecx|)} \overline{LR}(t) - \frac{t-1}{r}
\label{eq:RAL2}
\end{equation}
where $\tau(|\vecx|)$ denotes the cut-off step,
and $r=|y|/|x|$ is the target-to-source length ratio.

\subsection{Revision Rate}
\label{sec:revrate}

Since each modification on the target side would cost extra effort for the
audience to read,
we penalize all the revisions during the translation.
We define the revision rate as follows:
\vspace{-0.2cm}
\begin{equation*}
\vspace{-0.2cm}
\Big(\sum_{s=1}^{|x|-1} \text{dist} \Big(f(x_{\leqslant s}),f(x_{\leqslant s+1})\Big)\Big) \Big/ \Big(\sum_{s=1}^{|x|} |f(x_{\leqslant s})|\Big)
\label{eq:RR}
\end{equation*}
where $\text{dist}$ can be arbitrary distance measurement between two sequences.
For simplicity, we design a modified Hamming Distance to measure the difference:
\vspace{-0.2cm}
\begin{equation*}
\vspace{-0.2cm}
\text{dist} (a, b) = \text{hamming} \big(a, \ b_{\leq |a|} \circ \pad^{\max(|a|-|b|, 0)}\big)
\label{eq:hamming}
\end{equation*}
where \pad is a padding symbol in case $b$ is shorter than $a$.


\vspace{-6pt}
\begin{figure*}[!h]
\vspace{-6pt}
\begin{tabular}{cc}
\centering
  \includegraphics[width=7.3cm]{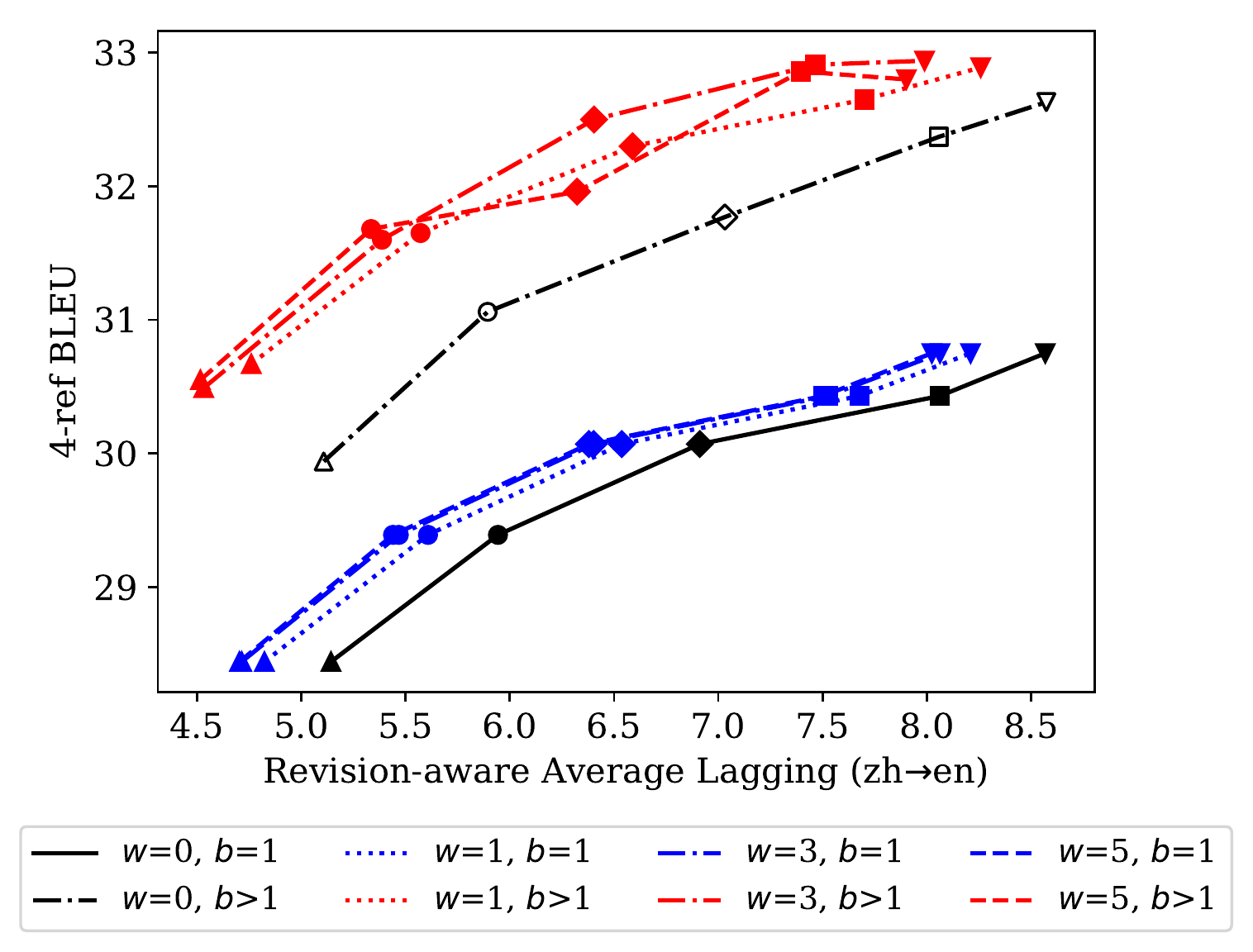} & \qquad
\centering
 \includegraphics[width=7.3cm]{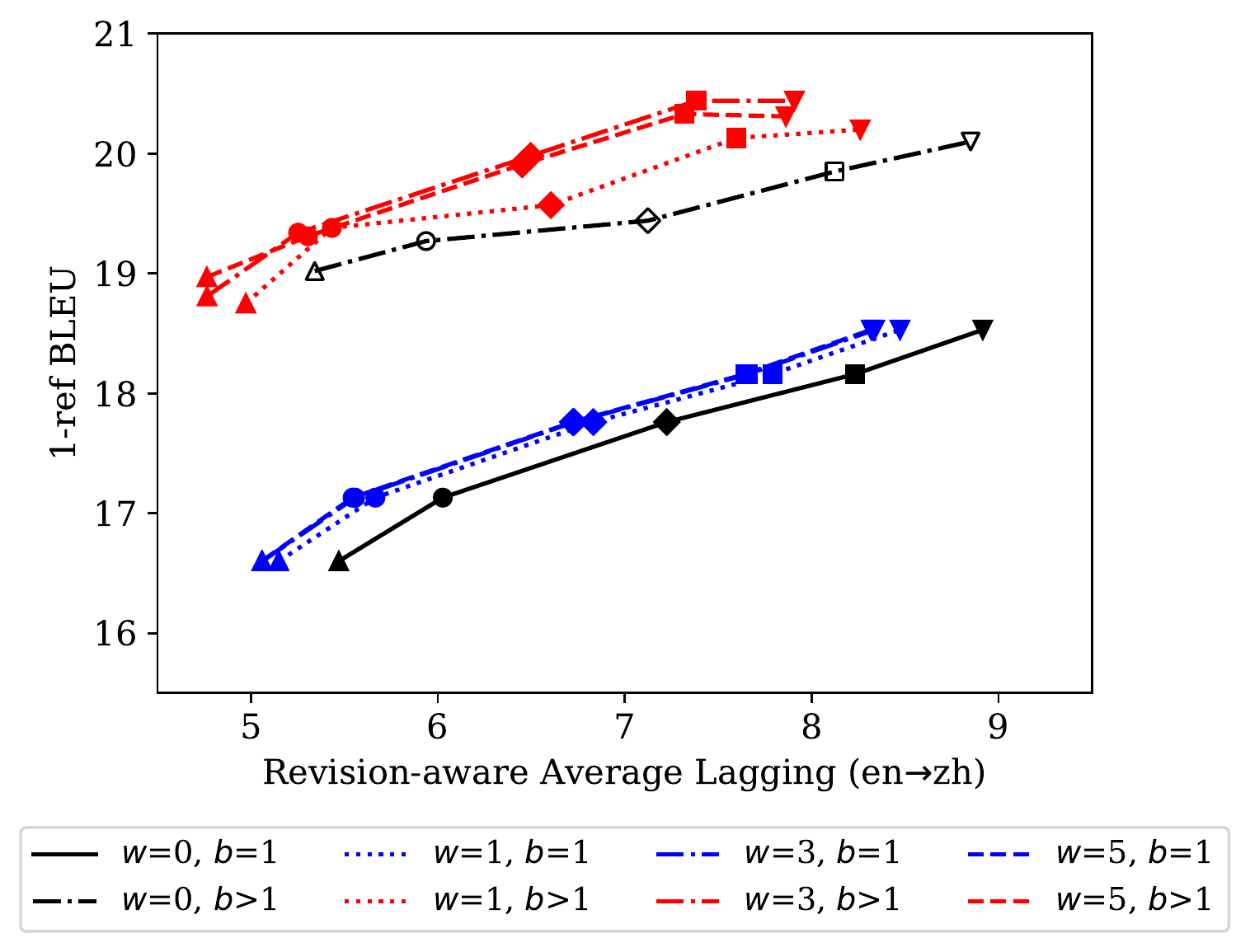}
\end{tabular}\\[-0.4cm]
\caption{
  BLEU against RAL 
  using adaptive policies.
Baseline is decoded with $w=0, b=1$ and $w=0, b>1$.
}
\label{fig:adapt_bleu}
\end{figure*}

\section{Experiments}


\paragraph{Datasets and Implementation}

We evaluate our work on Chinese-to-English and English-to-Chinese simultaneous translation tasks. 
We use the   
NIST corpus (2M sentence pairs) as the training data.
We first apply BPE~\cite{sennrich+:2015} on all texts 
to reduce the vocabulary sizes. 
For evaluation,
we use NIST 2006 and 
NIST 2008 as our dev and test sets 
with 4 English references.
We re-implement wait-$k$ model \cite{ma+:2019} 
and adaptive policy 
\cite{zheng+:2019b}.
We use Transformer \cite{vaswani+:2017} based
wait-$k$ model and 
pre-trained full-sentence model for learning adaptive policy.

\paragraph{Performance on Wait-$k$ Policy}

We perform experiments using opportunistic decoding on
wait-$k$ policies with $k\in \{1, 3, 5, 7, 9\}$,
opportunistic window $w\in \{1, 3, 5\}$
and beam size $b \in \{1, 3, 5, 7, 10, 15\}$.
We select the best beam size for each policy and 
window pair on dev-set.

We compare our proposed method with a baseline 
called re-translation which
uses a full-sentence NMT model to
re-decode the whole target sentence once a new
source word is observed.
The final output sentences of this method are identical
to the full sentence translation output with the same model
but the latency
is reduced.

Fig.~\ref{fig:waitk_bleu} (left) shows the 
Chinese-to-English
results of our proposed algorithm.
Since our greedy opportunistic decoding doesn't 
change the final output, there is no difference in BLEU
compared with normal decoding, but the latency is reduced.
However, by applying beam search,
we can achieve 3.1 BLEU improvement
and 2.4 latency reduction on wait-7 policy.

Fig.~\ref{fig:waitk_bleu} (right) shows the 
English-to-Chinese results.
Compare to the Chinese-to-English
translation results in previous section,
there is comparatively
less latency reduction by using beam search because
the output translations are slightly longer which
hurts the latency.
As shown in Fig.~\ref{fig:waitk_rate}(right),
the revision rate is still controlled under 8\%.

Fig.~\ref{fig:waitk_rate} shows the revision rate with different
window size on wait-$k$ policies.
In general, with opportunity window $w \le 5$, 
the revision rate of our proposed approach is under $8\%$,
which is much lower than re-translation.


\paragraph{Performance on Adaptive Policy}

Fig.~\ref{fig:adapt_bleu} shows the performance of the proposed
algorithm on adaptive policies.
We use threshold $\rho \in \{0.55, 0.53, 0.5, 0.47, 0.45 \}$.
We vary beam size $b \in \{1, 3, 5, 7, 10\}$ and select
the best one on dev-set.
Comparing with conventional beam search on consecutive writes,
our decoding algorithm achieves even much higher BLEU 
and less latency.


\subsection{Revision Rate vs. Window Size}

\begin{figure}[!h]
\vspace{-0.5cm}
\begin{tabular}{c}
\centering
  \includegraphics[width=6.9cm]{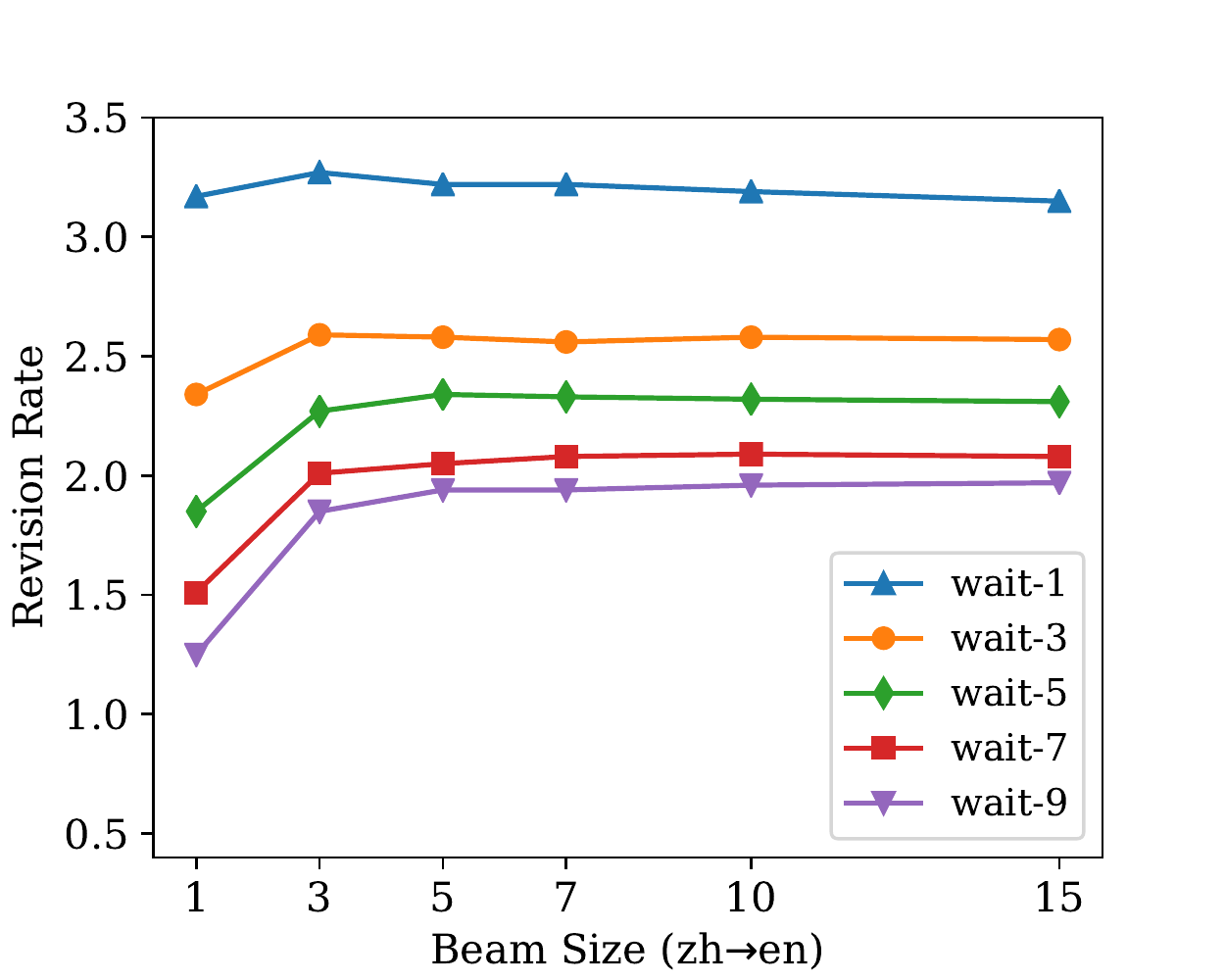}
\end{tabular}\\[-0.4cm]
\captionof{figure}{
Revision rate against beam size with window size of 3 and different
wait-$k$ policies.
}
\label{fig:beam_rate}
\end{figure}

We further investigate the revision rate with
different beam sizes on wait-$k$ policies.
Fig.~\ref{fig:beam_rate} shows that the revision rate
is higher with lower wait-$k$ policies.
This makes sense because the low $k$ policies are
always more aggressive and easy to make mistakes.
Moreover, we can find that the revision rate is
not very sensitive to beam size.

\section{Conclusions}
We have proposed an opportunistic decoding timely correction technique
which improves the latency and quality for simultaneous translation.
We also defined two metrics for revision-enabled simultaneous translation for the first time.

\section*{Acknowledgments}

L.~H.~was supported in part by NSF IIS-1817231.

\clearpage

\bibliography{main}
\bibliographystyle{acl_natbib}

\end{document}